%% file: main.tex
\documentclass[letterpaper]{article} 
\usepackage{aaai23}  
\usepackage{times}  
\usepackage{helvet}  
\usepackage{courier}  
\usepackage[hyphens]{url}  
\usepackage{graphicx} 
\usepackage{amsfonts}
\usepackage[table,xcdraw]{xcolor}
\usepackage{natbib}  
\usepackage{caption} 
\usepackage{todonotes}

\usepackage{booktabs}

\usepackage{algorithm}
\usepackage{algorithmic}
\newcommand{\ModelName}{\textsc{Lyra}}
\usepackage{newfloat}
\usepackage{listings}
\usepackage{multirow}
\usepackage{amsmath}
\usepackage{hyperref}
\usepackage[table,xcdraw]{xcolor}
\usepackage[normalem]{ulem}
\useunder{\uline}{\ul}{}
\DeclareCaptionStyle{ruled}{labelfont=normalfont,labelsep=colon,strut=off} 
\floatstyle{ruled}
\newfloat{listing}{tb}{lst}{}
\floatname{listing}{Listing}
\title{Unsupervised Melody-Guided Lyrics Generation
}
\author{
    Yufei Tian\textsuperscript{\rm 1}\thanks{Work was done when the author interned at Amazon.}, Anjali Narayan-Chen\textsuperscript{\rm 2}, Shereen Oraby\textsuperscript{\rm 2}, Alessandra Cervone\textsuperscript{\rm 2},
    Gunnar Sigurdsson\textsuperscript{\rm 2},\\ Chenyang Tao\textsuperscript{\rm 2}, Wenbo Zhao\textsuperscript{\rm 2}, Tagyoung Chung\textsuperscript{\rm 2}, Jing Huang\textsuperscript{\rm 2}, Nanyun Peng\textsuperscript{\rm 1,2}
}
\affiliations{
    \textsuperscript{\rm 1} University of California, Los Angeles, \textsuperscript{\rm 2} Amazon Alexa AI\\
    yufeit@g.ucla.edu, \{naraanja,orabys,cervon,gsig,chenyt,wenbzhao,tagyoung,jhuangz\}@amazon.com, violetpeng@cs.ucla.edu
}

\usepackage{bibentry}

\begin{document}

\maketitle

\begin{abstract}
Automatic song writing is a topic of significant practical interest. However, its research is largely hindered by the lack of training data due to copyright concerns and challenged by its creative nature.  
Most noticeably, prior works often fall short of modeling the cross-modal correlation between melody and lyrics due to limited parallel data, hence generating lyrics that are less singable.
Existing works also lack effective mechanisms for content control, a much desired feature for democratizing song creation for people with limited music background. 
In this work, we propose to generate pleasantly listenable 
lyrics without training on melody-lyric aligned data. Instead, we design a hierarchical lyric generation framework that disentangles training (based purely on text) from inference (melody-guided text generation). At inference time, we leverage the crucial alignments between melody and lyrics and compile the given melody into constraints to guide the generation process.
Evaluation results show that our model can generate high-quality lyrics that are more singable, intelligible, coherent, and in rhyme than strong baselines including those supervised on parallel data.
\end{abstract}

\input{01-intro}

\input{03-method}
\input{04-experiments}
\input{06-conclude}

\bibliography{aaai23}
\end{document}

%% file: 01-intro.tex
\begin{figure*}[t!]
  \centering
    \includegraphics[width=0.75\textwidth]{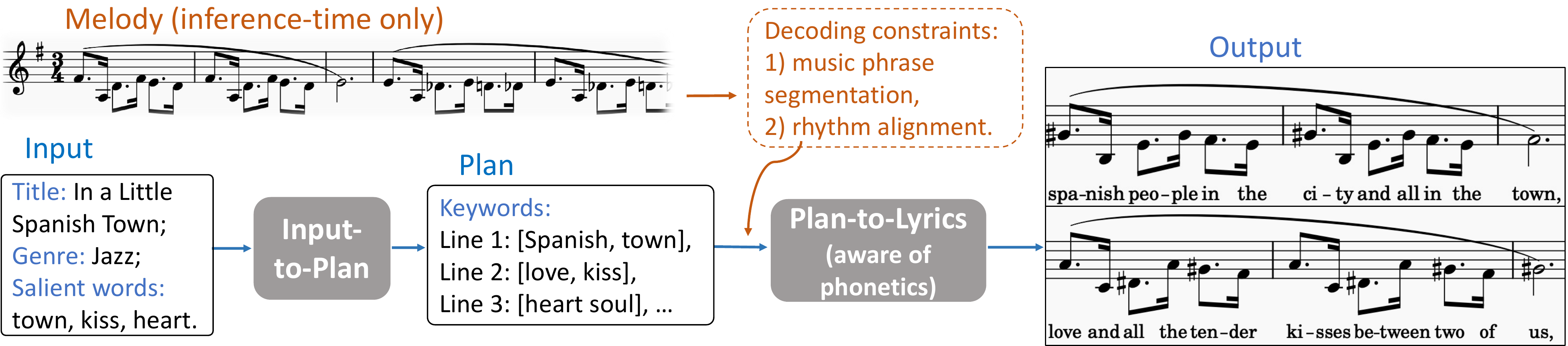}
    \vspace{-3 mm}
  \caption{ \small{An overview of our approach that disentangles training from inference. During training, our input-to-plan model learns to predict the sentence-level plan (i.e., keywords) given the title, genre, and salient words as input. Then, the plan-to-lyrics model generates the lyrics while being aware of word phonetic information and syllable counts. At inference time, we compile the given melody into 1) music phrase segments and 2) rhythm constraints to guide the generation. Examples of lyrics generated by the complete pipeline can be found \href{https://sites.google.com/view/lyricsgendemo}{here}.}}
  \label{fig:overview}    
  \vspace{-3 mm}
\end{figure*}
  \vspace{-3 mm}\section{Introduction}\label{sec:intro}

Music is ubiquitous and an indispensable part of humanity \cite{edensor2020national}. Self-served songwriting has thus become an emerging task and has attracted interest in the AI community \cite{watanabe2018melody,lee2019icomposer, yu2021conditional, sheng2021songmass, tan2021tutorial, zhang2022relyme}. However, the task of melody-to-lyric (M2L) generation, in which lyrics are generated based on a given melody,
faces three major challenges. First, there is a limited amount of melody-lyric aligned data. The process of collecting and annotating paired data is not only labor-intensive but also requires strong domain expertise and careful consideration of copyrighted source material.
Previous works either manually collect a small amount (usually a thousand) of melody-lyrics pairs \cite{watanabe2018melody, lee2019icomposer}, or \cite{sheng2021songmass} use the recently publicized data \cite{yu2021conditional} in which the lyrics are pre-tokenized at the syllable level leading to nonsensical syllables in the outputs. 

Another challenge lies in melody-lyric modeling. Compared with uni-modal sequence-to-sequence tasks such as machine translation, the latent correlation between lyrics and melody is difficult to learn. For example,
\citet{watanabe2018melody, lee2019icomposer, chen2020melody} train RNNs, LSTMs, or SeqGANs \cite{yu2017seqgan} where the input is a short melody piece and output is a line of lyrics; \citet{sheng2021songmass} train two transformers with separate encoder-decoders for each modality (lyrics or melody) and adopt cross attention \cite{vaswani2017attention} to learn the melody-lyrics mapping. However, these methods may generate less singable lyrics when they violate too often a superficial yet crucial alignment: one word in a lyric always strictly aligns with one or more notes in the melody \cite{nichols2009relationships}. In addition, their outputs are not fluent enough because training neural models from scratch cannot compete with massive pre-trained language models (PTLMs).

Besides, current M2L methods, all modelled as sequence-to-sequence, can only produce arbitrary lyrics based on the input melody sequence and have little control over the content to be generated \cite{chen2020melody, xue2021deeprapper, sheng2021songmass}. Such a problem setting is far away from real-world application scenarios where users will likely have an intended topic (e.g., an intended title and a few keywords) to write about. Hence, additional mechanisms to endow a lyric generator with topic relevance are urgently needed.

A promising direction to overcome the first two difficulties is to look into music theory studies, and pose melody alignment as decoding constraints for lyrics~\cite{guo-etal-2022-automatic}. One interesting discovery is that the \textit{durations} of music notes, not the pitch values, play a significant role in melody-lyric correlation~\cite{dzhambazov2017knowledge}. Intuitively, the segmentation of lyrics should match the segmentation of music phrases for breathability \cite{watanabe2018melody}. \citet{nichols2009relationships} also find that shorter note durations tend to associate with unstressed syllables, while longer durations tend to associate with stressed syllables. We find that, even when equipped with state-of-the-art neural architectures, existing lyric generators which are already supervised on melody-lyrics aligned data still fail to capture these simple yet fundamental rules. 

With all this in mind, we propose \textbf{\ModelName{}}, an unsupervised melody-conditioned {\it LYRics generAtor} that can generate high-quality lyrics related to a provided topic without training on melody-lyric data. To circumvent the shortage of paired data and comply with a given topic, we train a text-based hierarchical framework \citep{fan2018hierarchical, yao2019plan} that first predicts the content outline and then generates the lyrics.
At inference time, we compile a melody into constraints to guide the above text-based generation model. 
We summarize our contributions as follow:
\begin{itemize}
    \item To the best of our knowledge, we are among the first to generate melody-constrained lyrics without training on parallel data. To this end, we design a hierarchical framework that \textit{disentangles training from inference}.
    \item During training, our input-to-plan model learns to generate a plan of lyrics based on the input title and salient words, then the plan-to-lyrics model generates the complete lyrics. To fit in the characteristics of lyrics and melody, we equip the latter model with the ability to generate a sentence with a predefined count of syllables through multi-task learning. 
    \item Inspired by the aforementioned music theories, we design an inference-time decoding algorithm and incorporate two melody constraints: segment and rhythm alignment. Without losing flexibility, we introduce a factor to control the strength of the constraints. 
    \item Preliminary results show that our unsupervised model \textbf{\ModelName{}} can beat fully supervised baselines in terms of both text quality and musicality with a large margin.
\end{itemize}

%% file: 03-method.tex
\section{Methodology}
\subsection{Problem Setup}

We aim to generate lyrics that comply with both the provided topic and the melody. The input topic can be further decomposed into an intended title $T$ and a few salient words $S$ to be included in the generated lyrics. Following the settings of previous work \citep{chen2020melody, sheng2021songmass}, we also assume that the input melody $\mathcal{M}$ is predefined. $\mathcal{M}$ can be denoted by $\mathcal{M}=\{ p_1, p_2, ... p_M \}$, where each $p_i$ ($i \in {1,2, ...,M}$) is a music phrase. 
The music phrase can be further decomposed into timed music notes ($p_i = \{n_{i1}, n_{i2}, ... n_{iN_i}\}$), where each music note $n_{ij}$ ($j \in \{1,2, ...,N_i\}$) comes with a duration and is associated with or without a pitch value.\footnote{When a music note comes without a pitch value, it is a rest.} The output is a piece of lyrics $\mathcal{L}$ that aligns with the music notes: $\mathcal{L} = \{ w_{11}, w_{12}, ..., w_{MN}\}$. Here $ w_{ij}$ is a word or a syllable of a word that aligns with the music note $n_{ij}$.
\subsection{Training}
\paragraph{Overview and Data} An overview of our hierarchical generation method is shown in Figure \ref{fig:overview}. 
We finetune two modules in our purely text-based pipeline: 1)
an input-to-plan generator that generates a keyword-based intermediate plan, and 2) a plan-to-lyrics generator which is aware of word phonetics and syllable counts. Our training data consists of 38,000 English lyrics and their corresponding genres processed from the Lyrics for Genre Classification dataset.\footnote{\url{https://www.kaggle.com/datasets/mateibejan/multilingual-lyrics-for-genre-classification}}
\paragraph{Input-to-Plan}
The input contains the song title, the music genre, and three salient words extracted from ground truth lyrics using the YAKE algorithm \cite{campos2020yake}. Our input-to-plan model is then trained to generate a line-by-line keyword plan of the song.
Considering that in real time we might need different numbers of keywords for different expected output lengths, the number of planned keywords is not fixed. Specifically, we include a placeholder (the $<$MASK$>$ token) in the input for every keyword to be generated in the intermediate plan. In this way, we have control over how many keywords we would like per line. We finetune BART-large \cite{lewis2019bart} 
as our input-to-plan generator with format control.
\begin{table}[]
\centering
\resizebox{\columnwidth}{!}{%
\begin{tabular}{|c|c|c|}
\hline
Model                      &  (Input: syllable count) Output: generated lyric                \\ \hline
Ours, naïve                & (7) Cause the Christmas gift was for             \\ \hline
\citet{chen2020melody}    & (9) Hey now that's what you ever do my \\ \hline
\citet{sheng2021songmass} & (6) And then I saw you my                        \\ \hline
Ours, multitask & (12) Someday the tree is grown with other memories      \\ \hline
\end{tabular}
}
\caption{ \small {Examples of the desired syllable counts and the corresponding lyrics generated by different models. Our model with multitask learning is the only system that successfully generates a complete line of lyric. The supervised models \citep{chen2020melody, sheng2021songmass} which are already trained with the melody-lyrics paired data still generate dangling or cropped lyrics.}}
\label{table:dangling}    
\vspace{-5mm}
\end{table}

\paragraph{Plan-to-Lyrics} Our plan-to-lyrics module takes in the planned keywords as input and generates the lyrics. 
This module encounters an added challenge. To match the music notes of a given melody at inference time, this module should be capable of generating lyrics with a desired syllable count. If we naïvely force the generation to stop once it reaches the desired number of syllables, the outputs are usually cropped or dangling sentences. Surprisingly, existing lyric generators which are already trained on melody-to-lyrics aligned data also face the same issue. We list their example outputs of our naive model and two recent works in Table \ref{table:dangling}. We hence propose to study an under-explored task: generating a line of lyric that 1) has the desired number of syllables, and 2) is not cropped. To solve this, we propose to equip the plan-to-lyrics module with the word phonetics information and the ability to count syllables. To this end, we adopt multi-task learning to include the aforementioned external knowledge during training. Specifically, we study the collective effect of the following tasks: \begin{itemize}
    \item Task 1: Plan to lyrics generation
    \item Task 2: Syllable count: sentence → number of syllables
    \item Task 3: Plan to unnatural lyrics generation where we specify the syllable counts of each word in the output
    \item Task 4: Word to phoneme translation
    
\end{itemize}
We finetune the GPT-2 large model \cite{radford2019language} 
on different combinations of the four tasks, as our preliminary experimental results showed that BART could not learn these multi-tasks as well as GPT-2. We show our model's success rate of generating complete chunks of words with a predefined number of syllables in Section \ref{subsec:num_syllable}. 

\subsection{Inference}\label{sec:inf}
In this subsection, we discuss how to compile a given melody into constraints to guide the decoding. We start with the most straightforward constraints introduced in Section \ref{sec:intro}: 1) segmentation alignment and 2) rhythm alignment. Note that such melody constraints can be updated without needing to retrain the model.

\subsubsection{Segment Alignment}
The segmentation of music phrases should align with the segmentation of lyrics \cite{watanabe2018melody}. Given a melody, we first parse the melody into music phrases, then compute the number of music notes within each music phrase. For example, the first melody segment in Figure \ref{fig:overview} consists of 13 music notes, which should be equal to the number of syllables in the corresponding lyric chunk.  Without losing generality, we also add variations to this constraint where multiple notes can correspond to one single syllable when we observe such variations in the gold lyrics.

\subsubsection{Rhythm Alignment}
According to \citeauthor{nichols2009relationships}, the stress-duration alignment rule hypothesizes that music rhythm should align with lyrics meter. Namely, shorter note durations are more likely to be associated with unstressed syllables. At inference time, we `translate' a music note to a stressed syllable (denoted by 1)  or an unstressed syllable (denoted by 0) by comparing its duration to the average note duration. For example, based on the note durations, the first music phrase in Figure \ref{fig:overview} is translated into alternating 1s and 0s to guide the inference decoding.

\paragraph{Inference Decoding}
At each decoding step, we ask the plan-to-lyrics model to generate candidate complete words, instead of subwords, which is the default word piece unit for GPT-2 models. This enables us to retrieve the word phonemes from the CMU pronunciation dictionary \cite{weide1998carnegie} and identify the resulting syllable stresses. For example, since the phoneme of the word `Spanish' is `S PAE1 NIH0 SH', we can derive that it consists of 2 syllables that are stressed and unstressed. 

Next, we check if the candidate words satisfy the stress-duration alignment rule. Given a candidate word $w_i$ and the original logit $p(w_i)$ predicted by the plan-to-lyrics model, we introduce a factor $\alpha$ to control the strength:
\begin{equation}
    p'(w_i) =
    \begin{cases}
        p(w_i) & \text{if $w_i$ satisfies the rhythm alignment,}\\
        \alpha p(w_i) & \text{otherwise.}
    \end{cases}
\end{equation}
We can either impose a \textbf{hard constraint}, where we reject all those candidates that do not satisfy the rhythm rules ($\alpha$ = 0), or impose a \textbf{soft constraint}, where we would reduce their sampling probabilities ($0 < \alpha < 1$). 

%% file: 04-experiments.tex
\section{Experimental Results }
\subsection{Generating the correct number of syllables}\label{subsec:num_syllable}
Recall that we trained the plan-to-lyrics generator on four tasks in order to equip it with the ability to generate a sentence with a pre-defined number of syllables.
To test this feature, we computed the success rate on a held-out test set that contains 168 songs with 672 lines of lyrics. We experimented with both greedy decoding and sampling and report the result in Table \ref{table:syllable-count}. 
We can see that without multi-task learning, the model success rate is around 20\%, which is far from ideal. By gradually training with additional phonetics information, the success rate increases, reaching over 90\%. We also notice that the phoneme translation task is not helpful for our goal, so we disregard the last task and only keep the remaining three tasks in our final implementation.

\begin{table}[]
\centering
\small
\resizebox{.7\columnwidth}{!}{%
\begin{tabular}{cccccc}
\toprule
\multicolumn{4}{c}{Task Name} & \multicolumn{2}{c}{Success Rate} \\
\midrule
\multicolumn{1}{c}{\begin{tabular}[c]{@{}c@{}}T1\\ \scriptsize{Lyrics}\end{tabular}} & \multicolumn{1}{c}{\begin{tabular}[c]{@{}c@{}}T2\\ \scriptsize{Count}\end{tabular}} & \multicolumn{1}{c}{\begin{tabular}[c]{@{}c@{}}T3\\\scriptsize{Unnatural}\end{tabular}} & \begin{tabular}[c]{@{}c@{}}T4\\\scriptsize{Phoneme}\end{tabular} & \multicolumn{1}{c}{Greedy}  & Sample  \\ 
\midrule
\multicolumn{1}{c}{ $\checkmark$ }           & \multicolumn{1}{c}{}                                                                             & \multicolumn{1}{c}{}                                                                   &                                                                    & \multicolumn{1}{c}{23.14\%} & 19.87\% \\ 
\multicolumn{1}{c}{ $\checkmark$ }           & \multicolumn{1}{c}{$\checkmark$}                                                                             & \multicolumn{1}{c}{}                                                                   &  \multicolumn{1}{c}{}                                                                   & \multicolumn{1}{c}{50.14\%} & 44.64\% \\ 
\multicolumn{1}{c}{}            & \multicolumn{1}{c}{$\checkmark$}                                                                            & \multicolumn{1}{c}{$\checkmark$ }                                                                   &  \multicolumn{1}{c}{}                                                                   & \multicolumn{1}{c}{55.01\%} & 49.70\% \\ 
\multicolumn{1}{c}{ $\checkmark$ }           & \multicolumn{1}{c}{ $\checkmark$ }                                                                            & \multicolumn{1}{c}{$\checkmark$}                                                                   &  \multicolumn{1}{c}{}                                                                   & \multicolumn{1}{c}{\textbf{93.60\%}} & \textbf{89.13\% }\\ 
\multicolumn{1}{c}{ $\checkmark$ }           & \multicolumn{1}{c}{ $\checkmark$ }                                                                            & \multicolumn{1}{c}{ $\checkmark$ }                                                                  &  $\checkmark$                                                                   & \multicolumn{1}{c}{91.37\%} & 87.65\% \\ 
\bottomrule
\end{tabular}
}
\caption{Success rate for variants of our plan-to-lyrics model on generating sentences with desired number of syllables.}
\label{table:syllable-count}
\vspace{-5mm}
\end{table}

\begin{table*}[t!]\vspace{-2 mm}
\small
\centering
\begin{tabular}{|c|ccc|cc|c|cc|}
\hline
\rowcolor[HTML]{E7E6E6} 
\cellcolor[HTML]{E7E6E6}                             & \multicolumn{3}{c|}{\cellcolor[HTML]{E7E6E6}Topic Relevance}                                                                                                                                                                                                              & \multicolumn{2}{c|}{\cellcolor[HTML]{E7E6E6}Diversity}             & \multicolumn{2}{c|}{\cellcolor[HTML]{E7E6E6}Fluency}     & Music Alignment                                                              \\ \cline{2-9} 
\rowcolor[HTML]{E7E6E6} 
\multirow{-2}{*}{\cellcolor[HTML]{E7E6E6}Model Name} & \multicolumn{1}{l|}{\cellcolor[HTML]{E7E6E6}\begin{tabular}[c]{@{}l@{}}Salient Word\\ Coverage↑\end{tabular}} & \multicolumn{1}{c|}{\cellcolor[HTML]{E7E6E6}\begin{tabular}[c]{@{}c@{}}Sent \\ Bleu↑\end{tabular}} & \begin{tabular}[c]{@{}c@{}}Corpus  \\ Bleu↑\end{tabular} & \multicolumn{1}{c|}{\cellcolor[HTML]{E7E6E6}Dist-1↑} & Dist-2↑        & PPL ↓       & \multicolumn{1}{c|}{\cellcolor[HTML]{E7E6E6}\begin{tabular}[c]{@{}c@{}}Cropped \\ Sentence↓\end{tabular}} & 
\multicolumn{1}{c|}{\cellcolor[HTML]{E7E6E6}\begin{tabular}[c]{@{}c@{}}Stress-\\Duration\end{tabular}}        \\ \hline
SongMASS                                             & \multicolumn{1}{c|}{/}                                                                                       & \multicolumn{1}{c|}{0.045}                                                                        & 0.006                                                   & \multicolumn{1}{c|}{\textbf{0.17}}                  & \textbf{0.57} & 518         & \multicolumn{1}{c|}{34.51\%}                                                                            & 58.8\%          \\ \hline
GPT-2   finetuned                                    & \multicolumn{1}{c|}{/}                                                                                       & \multicolumn{1}{c|}{0.026}                                                                        & 0.020                                                   & \multicolumn{1}{c|}{0.09}                           & 0.31          & \textbf{82} & \multicolumn{1}{c|}{/}                                                                                  & 53.6\%          \\ \hline
\ModelName{} w/o rhythm                                      & \multicolumn{1}{c|}{\textbf{91.8\%}}                                                                         & \multicolumn{1}{c|}{{\ul 0.074}}                                                                  & {\ul 0.046}                                             & \multicolumn{1}{c|}{{\ul 0.12}}                     & 0.45          & {\ul 85}    & \multicolumn{1}{c|}{\textbf{3.65\%}}                                                                    & 63.1\%          \\ \hline
\ModelName{} w/   soft rhythm                                 & \multicolumn{1}{c|}{{\ul 89.4\%}}                                                                            & \multicolumn{1}{c|}{\textbf{0.075}}                                                               & \textbf{0.047}                                          & \multicolumn{1}{c|}{0.11}                           & {\ul 0.46}    & {\ul 85}    & \multicolumn{1}{c|}{{\ul 8.96\%}}                                                                       & {\ul 68.4\%}    \\ \hline
\ModelName{} w/ hard rhythm                                 & \multicolumn{1}{c|}{88.7\%}                                                                                  & \multicolumn{1}{c|}{0.071}                                                                        & 0.042                                                   & \multicolumn{1}{c|}{{\ul 0.12}}                     & 0.45          & 108         & \multicolumn{1}{c|}{10.26\%}                                                                            & \textbf{89.5\%} \\ \hline
\rowcolor[HTML]{E7E6E6} 
Ground   Truth                                       & \multicolumn{1}{c|}{\cellcolor[HTML]{E7E6E6}100\%}                                                           & \multicolumn{1}{c|}{\cellcolor[HTML]{E7E6E6}1.000}                                                & 1.000                                                   & \multicolumn{1}{c|}{\cellcolor[HTML]{E7E6E6}0.14}   & 0.58          & 93          & \multicolumn{1}{c|}{\cellcolor[HTML]{E7E6E6}3.92\%}                                                     & 73.3\%          \\ \hline
\end{tabular}\vspace{-2.7 mm}
\caption{\small{Automatic evaluation results on 120 songs. The gold performance is highlighted in a grey background. Among all machine performances, we highlight the best scores in boldface and underline the second best.}\vspace{-2 mm}
  }
  \label{table:auto_eval}  
\end{table*}
\begin{figure*}[!t]
  \centering
  \vspace{-2.7 mm}
    \includegraphics[width=0.75\textwidth]{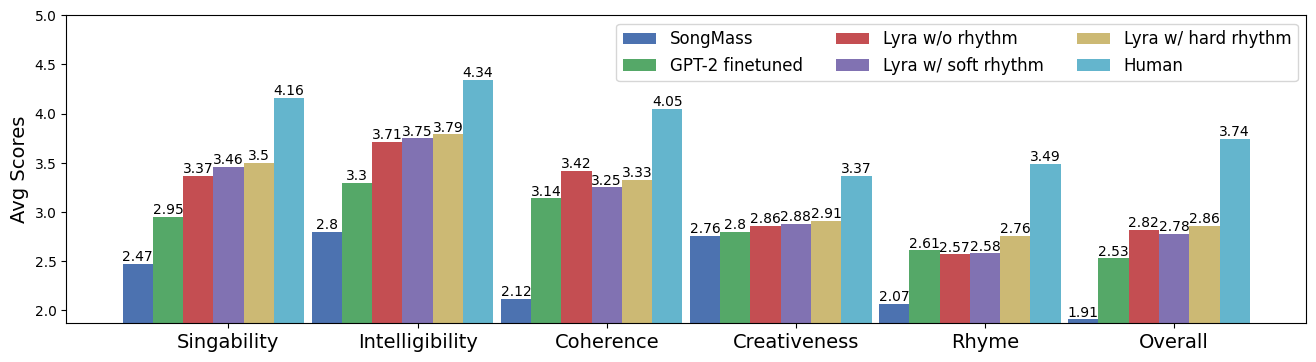}
  \vspace{-3.7 mm}
  \caption{\small{Average human annotator Likert scores for 5 songs on musicality, quality of lyrics alone, and overall. Considering the small sample size, the current results may be inaccurate for the comparison between variations of our own \ModelName{} models, as sometimes the differences are not significant enough. We plan to launch a large-scale human evaluation after recruiting qualified annotators to reduce the randomness.}}
  \label{fig:human_eval}    
  \vspace{-4 mm}
\end{figure*}
\subsection{Baseline Models for Lyrics Generation}
We compared the following models. \textbf{1. SongMASS} \cite{sheng2021songmass} is a state-of-the-art song writing system which leverages masked sequence to sequence pre-training and attention based alignment for M2L generation. It requires melody-lyrics aligned training data while our model does not. \textbf{2. GPT-2 finetuned on lyrics} is a uni-modal, melody-unaware GPT-2 large model that is finetuned end-to-end on the same lyrics data as our model. The input prompt contains a song title but not salient words. 
This serves as ablation of \textbf{\ModelName{} w/o rhythm} to test the efficacy of our plan-and-write pipeline without inference-time constraints.
\textbf{3. \ModelName{} w/o rhythm} is our base model consisting of the input-to-plan and plan-to-lyrics modules with syllable control, but without the rhythm alignment.
\textbf{4. \ModelName{} w/ (soft/hard) rhythm} is our multi-modal model with music segmentation and soft or hard rhythm constraints. For the soft constraints setting, the strength controlling hyperparameter $\alpha=0.01$.

\subsection{Automatic Evaluation}
We automatically assessed the generated lyrics on two aspects: the quality of text, and music alignment. For \textbf{text quality}, we divided it into 3 aspects: 1) \textbf{Topic Relevance}, measured by input salient word coverage ratio, and sentence- or corpus-level BLEU \cite{papineni2002bleu}; 2) \textbf{Diversity}, measured by distinct unigrams and bigrams \cite{li2015diversity}; 3) \textbf{Fluency}, measured by the perplexity computed using Huggingface's pretrained GPT-2 small. We also computed the ratio of cropped sentences among all sentences to assess how well they fit music phrase segments. For \textbf{music alignment}, we computed the percentage where the stress-duration rule holds.

We report the results in Table \ref{table:auto_eval}. Our \ModelName{} models significantly outperform the baselines and generate the most on-topic and fluent lyrics. In addition, adding rhythm constraints to the base \ModelName{} noticeably increases the music alignment quality without sacrificing too much text quality.
It is also noteworthy that humans do not consistently follow  stress-duration alignment, meaning that higher is not necessarily better for music alignment percentage.
Since the baseline model SongMASS has no control over the content, it has lowest topic relevance scores. Moreover, although the SongMASS baseline seems to achieve the best diversity, it tends to produce non-sensical sentences that consist of a few gibberish words (e.g. `for hanwn to stay with him when, he got to faney he alone'). Such degeneration is also reflected by the extremely high perplexity and cropped sentence ratio.

\subsection{Human Study}
We conducted a human evaluation on 5 randomly selected songs. We first used an online \href{http://www.sinsy.jp/}{singing voice synthesizer} \cite{hono2021sinsy}
to generate the singing voice audio clips. We then asked 20 Mechanical Turkers to evaluate the lyrics on six dimensions on a scale from 1 to 5, where a higher score means better quality. For musicality, we asked them to rate singability (whether the melody's rhythm aligns well with the lyric's rhythm)  and intelligibility (whether the content of the lyrics is easy to understand without looking at the lyrics). For the lyric quality, we asked them to rate coherence, creativeness, and rhyme. Finally, we asked annotators to rate how much they like the song overall. The average inter annotator agreement (Spearman correlation) is 0.61. 
The preliminary results are shown in Figure \ref{fig:human_eval}. 

Generated examples can be found in this \href{https://sites.google.com/view/lyricsgendemo}{anonymous demo website}. Surprisingly, SongMASS performs even worse than the finetuned GPT-2 baseline in terms of musicality. Upon further inspection, we posit that SongMASS too often deviates from common singing habits: it either assigns two or more syllables to one music note, or matches one syllable with three or more consecutive music notes. It is clear that \ModelName{} w/o rhythm greatly outperforms both baseline models in terms of singability, intelligibility, coherence, and overall quality, which demonstrates the efficacy of our plan-and-write pipeline with syllable control. The addition of rhythm alignment leads to further improvements in musicality, creativeness, and rhyme, with some sacrifices in coherence and overall quality.
Considering the small sample size, it is still early to draw a final conclusion. We plan to launch a large-scale human evaluation after recruiting qualified annotators for greater statistical significance.

%% file: 06-conclude.tex
\vspace{-2.7 mm}\section{Conclusion}

Our work explores the potential of lyrics generation without training on lyrics-melody aligned data.
To this end, we design a hierarchical plan-and-write framework that disentangles training from inference. 
At inference time, we compile the given melody into 1) music phrase segments, and 2) rhythm constraints. Such melody constraints can be updated without needing to retrain the model. 
Our preliminary results show that our model can generate high-quality lyrics that are singable, intelligible, and coherent.

\section*{Acknowledgements}
The authors would like to thank Yiwen Chen for helping to design and providing valuable insights to the human evaluation. We also thank the anonymous reviewers for the helpful comments.